\documentclass{article}

\usepackage{arxiv}

\usepackage[utf8]{inputenc} % allow utf-8 input
\usepackage[T1]{fontenc}    % use 8-bit T1 fonts
\usepackage{hyperref}       % hyperlinks
\usepackage{url}            % simple URL typesetting
\usepackage{booktabs}       % professional-quality tables
\usepackage{amsfonts}       % blackboard math symbols
\usepackage{nicefrac}       % compact symbols for 1/2, etc.
\usepackage{microtype}      % microtypography
\usepackage{lipsum}		% Can be removed after putting your text content
\usepackage{graphicx}
\usepackage[round]{natbib}
\usepackage{doi}
\usepackage{microtype}
\usepackage{graphicx}
\usepackage{subfigure}
\usepackage{booktabs}
\usepackage{algorithm}
\usepackage{algorithmic}
\usepackage{amsmath}
\usepackage{amssymb}
\usepackage{mathtools}
\usepackage{amsthm}

\title{MISLEAD: Manipulating Importance of Selected features for Learning Epsilon in Evasion Attack Deception}

\author{ \href{https://orcid.org/0009-0007-0223-3843}{\includegraphics[scale=0.06]{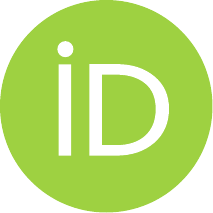}\hspace{1mm}Vidit Khazanchi} \\
	Department of Materials\\ 
	Indian Institute Technology\\
	Bombay, India\\
	\texttt{viditk0812@gmail.com} \\
	\And
	\href{https://orcid.org/0000-0002-8458-6795}{\includegraphics[scale=0.06]{orcid.pdf}\hspace{1mm}Pavan Kulkarni} \\
	AIShield \\ 
        Bosch Global Software Technologies \\
        Bangalore, India \\
	\texttt{pavan.kulkarni@in.bosch.com} \\
 \And
	\href{https://orcid.org/0000-0002-4247-4410}{\includegraphics[scale=0.06]{orcid.pdf}\hspace{1mm}Yuvaraj Govindarajulu} \\
	AIShield \\ 
        Bosch Global Software Technologies \\
        Bangalore, India \\
	\texttt{govindarajulu.yuvaraj@de.bosch.com} \\
 \And
	\href{https://orcid.org/0000-0002-1183-4399}{\includegraphics[scale=0.06]{orcid.pdf}\hspace{1mm}Manojkumar Parmar} \\
	AIShield \\ 
        Bosch Global Software Technologies \\
        Bangalore, India \\
	\texttt{manojkumar.parmar@in.bosch.com} \\
}

\renewcommand{\shorttitle}{\textit{MISLEAD}}

\begin{document}
\maketitle

\begin{abstract}
Emerging vulnerabilities in machine learning (ML) models due to adversarial attacks raise concerns about their reliability. Specifically, evasion attacks manipulate models by introducing precise perturbations to input data, causing erroneous predictions. To address this, we propose a methodology combining SHapley Additive exPlanations (SHAP) for feature importance analysis with an innovative Optimal Epsilon technique for conducting evasion attacks. Our approach begins with SHAP-based analysis to understand model vulnerabilities, crucial for devising targeted evasion strategies. The Optimal Epsilon technique, employing a Binary Search algorithm, efficiently determines the minimum epsilon needed for successful evasion. Evaluation across diverse machine learning architectures demonstrates the technique's precision in generating adversarial samples, underscoring its efficacy in manipulating model outcomes. This study emphasizes the critical importance of continuous assessment and monitoring to identify and mitigate potential security risks in machine learning systems.
\end{abstract}


\section{Introduction}
\label{sec:Intro}
The widespread adoption of machine learning models has driven remarkable technological advancements and improvements in decision-making, concurrently exposing a vulnerability—adversarial attacks, particularly evasion attacks. These attacks involve subtle alterations to input data, leading to erroneous predictions with potentially severe consequences \cite{szegedy2014intriguing, Goodfellow2014ExplainingAH, carlini7958570}. To address this challenge, our approach introduces a methodology combining SHapley Additive exPlanations (SHAP) for feature importance analysis with an innovative optimal epsilon technique \cite{NIPS2017_7062}. Motivated by the growing need to secure machine learning models in vital sectors like healthcare, finance, autonomous vehicles, and cybersecurity \cite{Biggio_2018}, our methodology integrates feature importance analysis using SHAP, a powerful tool for understanding feature impact across various domains \cite{articleLundbergscott}. This analysis spans both binary and multiclass classification scenarios, offering insights for developing targeted evasion strategies by evaluating the significance of different features.

The optimal epsilon technique, introduced in our study, plays a pivotal role in evasion attacks by determining the minimal epsilon necessary for successful evasion. This concept involves finding the smallest perturbation magnitude to deceive a machine learning model into making incorrect predictions without detectable alterations. The technique enhances precision and effectiveness in exposing vulnerabilities, underscoring the need for robust countermeasures \cite{Papernot2016PracticalBA}. Through experiments, we assess the methodology's effectiveness across diverse machine learning architectures and datasets, showcasing its ability to generate precise adversarial samples \cite{Wang_2021_CVPR}.

% \textcolor{red}{rewrite and use the content from the rebuttal submission (keep the references)}The contributions of this paper significantly impact the dynamic landscape of machine learning security and adversarial attacks \cite{huang2017adversarial}. Integrating feature importance analysis with evasion attacks enhances understanding of model vulnerabilities and attack precision \cite{Dvijotham2018TrainingVL}. The optimal epsilon technique refines adversarial sample creation, offering a more secure path in deploying machine learning models across critical applications \cite{pmlr-v80-athalye18b}. Emphasizing both technical innovation and practical applicability positions this research as a substantial contribution to the field.

The contributions of this paper are as follows:
\begin{itemize}
   \item \textbf{Integration of SHAP with Evasion Attacks:} The novel approach of systematically integrating SHAP-based feature importance analysis into the evasion attack process, allowing for targeted manipulation of the most influential features, leading to more efficient and effective attacks \cite{huang2017adversarial, Dvijotham2018TrainingVL}.

   \item \textbf{Optimal Epsilon Technique:} Introduction of a novel and systematic technique for determining the minimum epsilon needed for successful evasion through a binary search-based approach, enhancing the precision of adversarial sample generation and providing a nuanced understanding of model robustness \cite{pmlr-v80-athalye18b}.

   \item \textbf{Black-Box Applicability:} MISLEAD operates in a black-box setting, relying solely on the model's predictions, making it applicable to real-world scenarios where attackers might not have access to the model's internal parameters.

   \item \textbf{Comprehensive Feature Analysis:} Thorough analysis of feature impacts, categorizing them based on their influence and directionality, allowing for the development of sophisticated and targeted attack strategies.
\end{itemize}

The paper is organized as follows: Section \ref{sec:Background} explores fundamental theories and previous studies, providing a base for our research approach. In Section 3, we detail our methodology, including SHAP-based feature importance analysis and the innovative optimal epsilon technique for evasion attacks. Section 4 discusses our experimental setup and findings, highlighting the effectiveness and implications of our work. Finally, Section 5 concludes the paper with a summary of our findings and potential avenues for future research.

\section{Background}
\label{sec:Background}

\subsection{Adversarial Machine Learning}
Recent years have underscored machine learning models' vulnerability to adversarial attacks, especially evasion attacks that involve crafting adversarial examples for specific target class predictions. Adversarial attacks categorize based on the attacker’s knowledge: perfect (white-box), limited (gray-box), and zero knowledge (black-box) attacks \cite{nazemi2019potential, information_leakage, sotgiu2020deep, Biggio_2018}. These categories depend on the attacker's understanding of training data and model parameters.

\subsection{Feature Importance}
SHAP has emerged as a powerful tool for understanding machine learning models' decision-making process \cite{NIPS2017_7062}. Applied across domains, including image classification, natural language processing, and tabular data analysis \cite{9265985, 9904989, mosca-etal-2022-shap, lundberg2020local2global}, SHAP values enhance interpretability, aiding feature selection and model optimization \cite{CAI201870, Chen2018LearningTE, Ancona2017TowardsBU}. The optimal epsilon technique in our paper systematically determines the minimum epsilon for effective evasion, a valuable contribution to the field. While epsilon's role in controlling perturbation magnitude is well-discussed, systematic techniques for determining optimal epsilon are underexplored. Our paper adapts binary search algorithms, a novel approach in evasion attacks \cite{YU2017689, Han2012/09, articleMeyers}.

\subsection{Evasion attacks on ML models}
Fast Gradient Sign Method (FGSM) is one of the pioneering techniques in evasion attacks. It computes gradients with respect to the input data and perturbs the data in the direction that maximizes the loss, thus causing misclassification \cite{Goodfellow2014ExplainingAH}. Projected Gradient Descent (PGD) is an iterative variant of FGSM that performs multiple steps of gradient descent while ensuring that the perturbed data remains within an epsilon ball around the original sample \cite{articleMadry}. DeepFool is an attack method that computes the perturbation by linearizing the decision boundary of the model and iteratively finding the closest decision boundary point \cite{7780651}. Papernot et. al. have explored practical black-box attacks against machine learning models, emphasizing the real-world applicability of adversarial attacks \cite{Papernot2016PracticalBA}. Athalye et. al. investigated techniques for synthesizing robust adversarial examples, aiming to create adversarial samples that are less susceptible to detection and defense mechanisms \cite{pmlr-v80-athalye18b}.

\subsection{Feature + Evasion on Tabular classification}
In tabular datasets, attacks focus on domain-specific challenges, like financial datasets \cite{hashemi2020permuteattack, sarkar2018robust, cartella2021adversarial}. Novel methods, such as Max Salience Attack (MSA), aim to minimize altered features \cite{sarkar2018robust}. Our paper proposes a unique methodology integrating SHAP-based feature importance analysis into evasion attacks, providing a comprehensive perspective on model vulnerabilities. This differs from recent advancements like Feature Importance Guided Attack (FIGA) \cite{gressel2023feature}, emphasizing minimal perturbation using SHAP and an optimal epsilon technique.

\subsection{SHAP for Black Box Access}
Hassija et al., \cite{Hassija2024} explains, SHAP operates as a black-box explainer for black-box models:

\begin{itemize}
   \item \textbf{SHAP's Functionality:} SHAP focuses on explaining individual predictions, not the entire inner workings of the model.
   
   \item \textbf{Model Agnostic:} The key strength of SHAP lies in its model-agnostic nature. It doesn't require knowledge of the model's architecture (e.g., decision trees, neural networks) to compute feature contributions. It treats the model as a function, taking inputs and generating outputs.
   
   \item \textbf{SHAP's Internal Workings:} SHAP leverages game theory concepts (Shapley Values) to fairly distribute credit for a prediction amongst all features. While the underlying calculations involve the model's predictions for various data permutations, SHAP itself remains agnostic to the specific model logic.
\end{itemize}

In essence, SHAP acts as an intermediary. It interacts with the black-box model at the input-output level, extracting feature importance without needing to delve into the internal complexities of the model. This allows SHAP to provide valuable insights into a model's decision-making process without requiring white-box access.

\section{Methodology}
\label{sec:Methodology}
This section provides a detailed explanation of our evasion attack methodology, covering the overall threat model, key assumptions, data collection, and preprocessing procedures. The goal is to enhance the reliability and quality of our input data, emphasizing factors influencing the model's predictions through SHAP techniques for feature importance analysis. Following this analysis, we categorize the impact of individual features based on our findings and introduce an attack strategy, a carefully designed plan to systematically modify input samples, outsmarting the model's predictions to achieve the desired target class.

\subsection{Threat Model and Assumptions}

Our research assumes the existence of a machine learning target model vulnerable to evasion attacks due to its sensitivity to changes in input features. Focusing on tabular datasets with numerical and categorical data, our attack strategy operates in a complete black-box setting, where the attacker can only query the target model for predictions without insight into internal parameters and weights.

\subsection{Data Collection and Preprocessing}

We employ a Bank Marketing dataset comprising numerical and categorical variables. The class distribution shows 36,548 samples in Class 0 and 4,640 samples in Class 1. With 10 numerical and 10 categorical features, the dataset is comprehensively characterized. Categorical variables are numerically encoded using LabelEncoder \cite{scikit-learn}, and both numerical and categorical features are normalized to a 0 to 1 range using MinMaxScaler \cite{scikit-learn} for comparative analysis.

\subsection{Feature Importance Analysis using SHAP}

SHAP values quantify the influence of each feature on model predictions, providing insights into the direction and magnitude of their impact. Calculating SHAP values for every sample generates an array of values for the 20 features. These insights are analyzed through various plots to rank features, assess their average impact, and identify the most influential ones in the dataset. The application of SHAP is explored in both binary and multiclass classification scenarios.

\begin{figure}[t]
    \centering
        \subfigure[Global Bar Plot]{\includegraphics[width=3in]{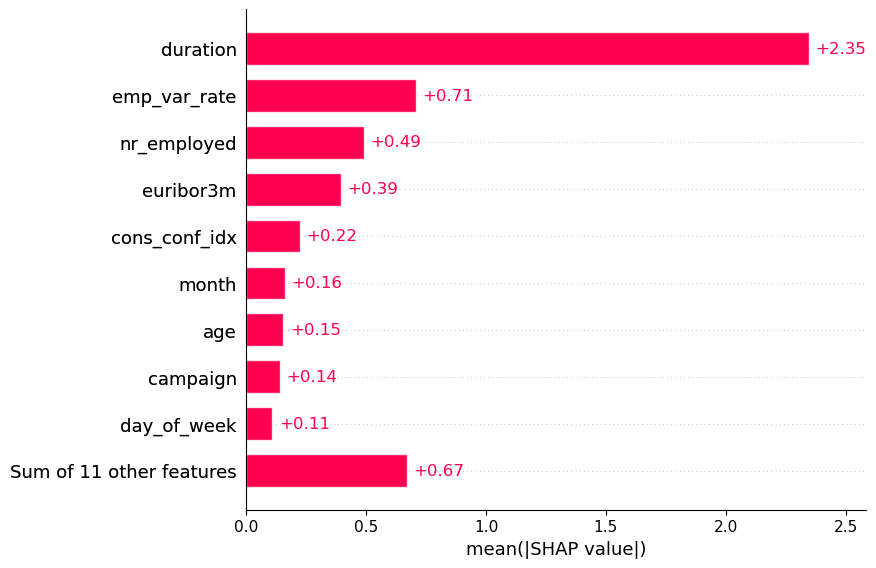}%
        \label{fig:bcglobal}}
        \hfil
        \subfigure[Local Bar Plot]{\includegraphics[width=3in]{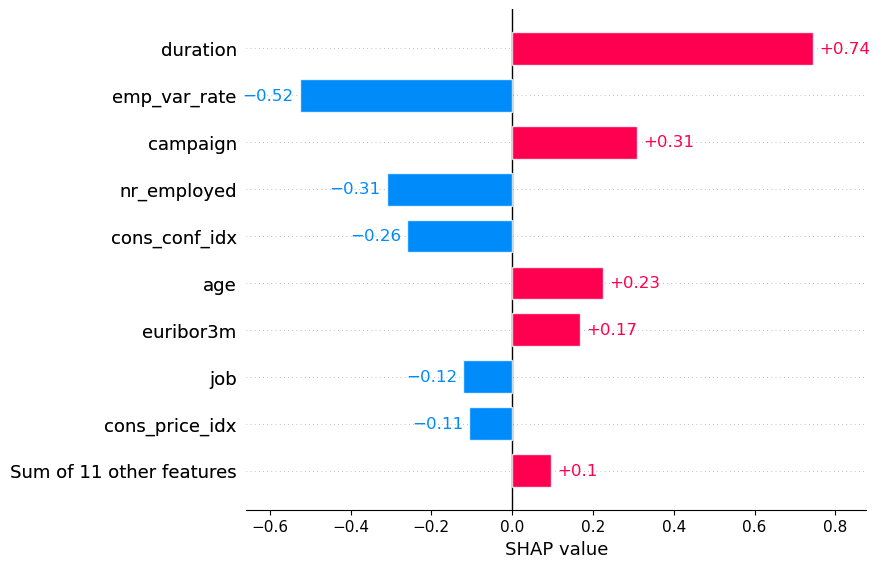}%
        \label{fig:bclocal}}
    \caption{Binary Classification Bar Plots}
    \label{fig:bcplots}
\end{figure}

\subsubsection{Binary Classification}

In binary classification, involving only two labels, interpreting SHAP values is more straightforward. The following plots are employed:

\textbf{Global Bar Plot:} Offers a comprehensive view of feature importance across the entire dataset. It displays the mean SHAP value for each feature, arranged in descending order of importance, highlighting influential features in driving model predictions (Figure \ref{fig:bcglobal}). However, it doesn't indicate the direction of impact or specific feature values necessary for predicting specific classes.

\textbf{Local Bar Plot:} Zooms in on individual samples, detailing how each feature impacts the model’s prediction for a specific sample. This plot enhances understanding at the micro-level, revealing intricate relationships between features and predictions (Figure \ref{fig:bclocal}).

\begin{figure}[t]
    \centering
    \includegraphics[width=3in]{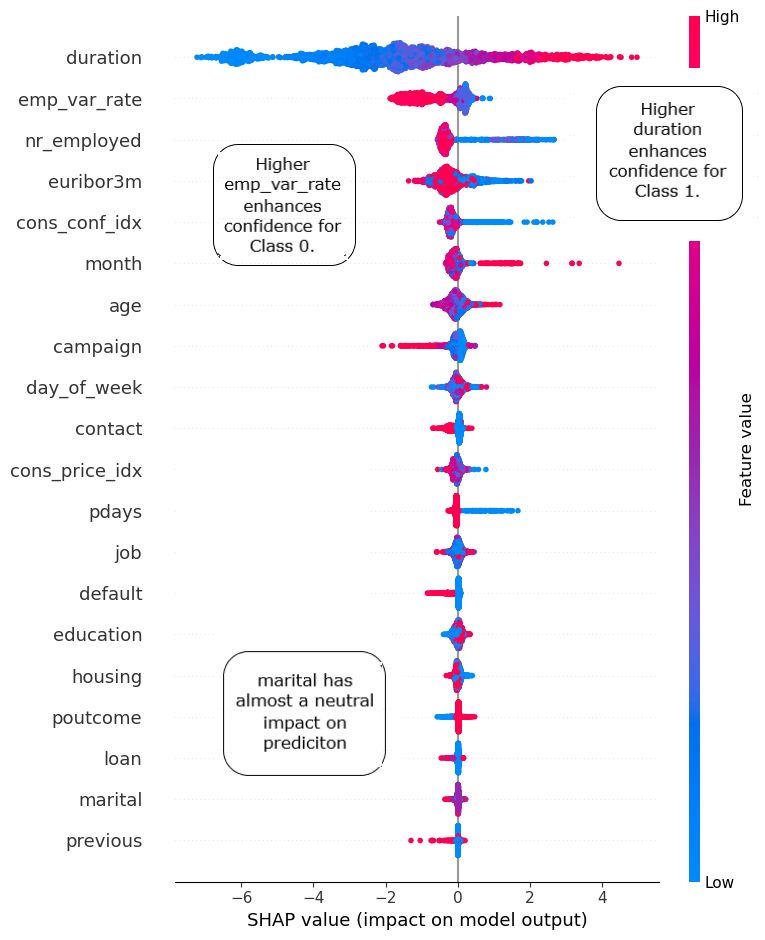}
    \caption{Binary Classification Beeswarm Plot}
    \label{fig:bcbeeswarm}
\end{figure}

\textbf{Beeswarm Plot:} Figure \ref{fig:bcbeeswarm} provides a more detailed and informative visualization than the bar plots. It showcases the relative importance of features and their relationship with the predicted outcome, offering a comprehensive overview of how variables influence predictions. This insight is critical for generating perturbations in our evasion attack strategy.

Together, these plots form an integral part of our methodology, enabling analysis of the impact of individual features on model predictions in both macro and micro perspectives.

\subsubsection{Multiclass Classification}

\begin{figure*}[t]
\centering
\subfigure[MultiClass Global Bar  Plot]{\includegraphics[width=3in]{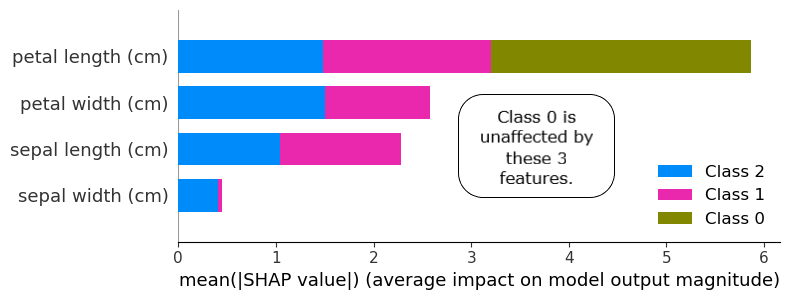}%
\label{fig:mcplots}}
\hfil
\subfigure[Beeswarm Plot For Class 0]{\includegraphics[width=3in]{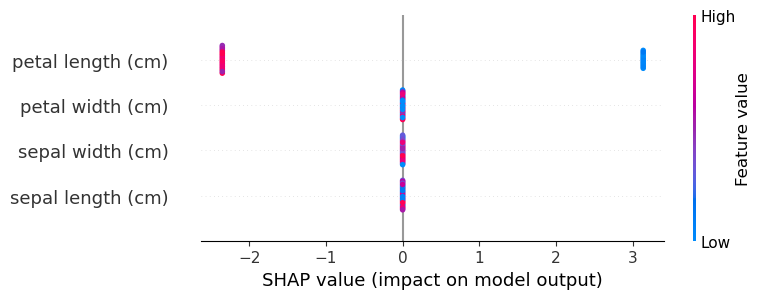}%
\label{fig:mcbee0}}
\hfil
\subfigure[Beeswarm Plot For Class 1]{\includegraphics[width=3in]{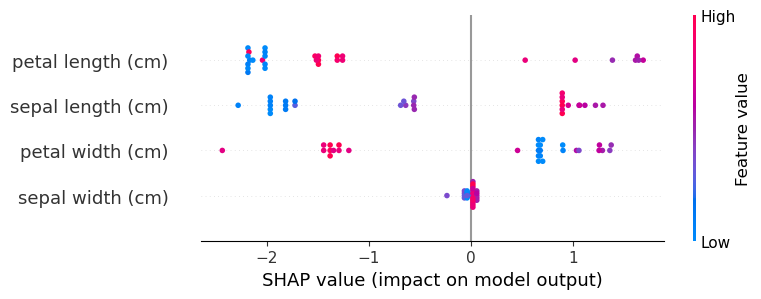}%
\label{fig:mcbee1}}
\hfil
\subfigure[Beeswarm Plot For Class 2]{\includegraphics[width=3in]{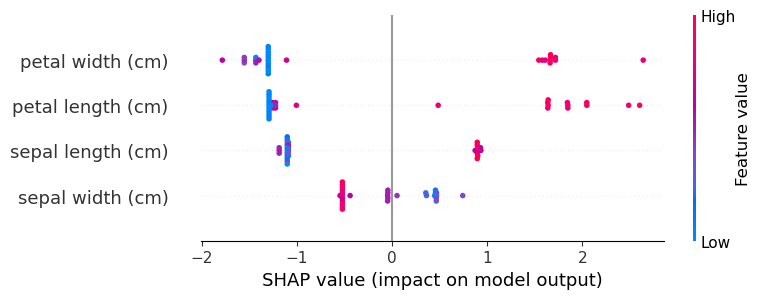}%
\label{fig:mcbee2}}

\caption{MutliClass Classification Beeswarm and Global Bar Plots}
\label{fig:mcbeeswarm}
\end{figure*}

Multiclass classification introduces more complexity with multiple classes for prediction. The following plots are applied:

\textbf{Global Bar Plot:} Represents each feature with a bar divided into sections corresponding to each class (Figure \ref{fig:mcplots}). This allows for a detailed understanding of a feature's importance across different classes, revealing the variable significance of features in predicting various classes.

\textbf{Beeswarm Plot:} Similar to the binary classification scenario, the multiclass beeswarm plot provides an intricate visualization of SHAP values across the dataset. However, a notable difference is the existence of separate graphs for each class (Figures \ref{fig:mcbee0}, \ref{fig:mcbee1}, \ref{fig:mcbee2}), illustrating how features influence predictions for a specific class. This detailed representation is key to understanding the optimal feature values necessary for predicting each class, crucial for devising targeted evasion strategies.

Together, these plots in multiclass classification scenarios enable a comprehensive analysis of feature impacts, both across and within individual classes. They enhance understanding of how different features contribute to model predictions in multiclass settings, vital for developing sophisticated evasion techniques in a multiclass context.

\subsection{Feature Analysis for Evasion}
In addressing the challenge of evasion attacks, we conduct a comprehensive analysis of feature impacts on model predictions. This section details our approach to categorize and utilize feature behavior for devising evasion strategies. By understanding how individual features influence model predictions, we aim to identify vulnerabilities in machine learning models and exploit these for successful evasion.

\subsubsection{Feature Impact Categorization}
Our methodology begins with categorizing the impacts of individual features, pivotal for understanding their influence on model predictions. We use predefined thresholds, denoted as \(T_{\text{low}}\) and \(T_{\text{high}}\), to assign impact categories: 'Low' (L), 'Medium' (M), or 'High' (H), based on the feature's value \(F_{ij}\) using Equation \ref{eq:festCat}.

\begin{equation}
CF_{ij} = \begin{cases} 
     L & \text{if } F_{ij} < T_{low} \\
     M & \text{if } T_{low} \leq F_{ij} < T_{high} \\
     H & \text{if } F_{ij} \geq T_{high} 
   \end{cases}
% g(t)=t^{N-1}e^{-\alpha t}\cos(\omega_{o}t)u(t),
\label{eq:festCat}
\end{equation}

where, \(CF_{ij}\) represents the impact category of feature \(i\) in sample \(j\)

\subsubsection{Categorizing SHAP Values}

Additionally, we categorize the SHAP values \(S_{ij}\), labeling them as 'positive' (P), 'neutral' (\(N_T\)), or 'negative' (N), based on their sign using Equation \ref{eq:shapCat}.

\begin{equation}
     CS_{ij} = \begin{cases} 
     P & \text{if } S_{ij} > 0 \\
     N_{T} & \text{if } S_{ij} = 0 \\
     N & \text{if } S_{ij} < 0 
   \end{cases}
% g(t)=t^{N-1}e^{-\alpha t}\cos(\omega_{o}t)u(t),
\label{eq:shapCat}
\end{equation}

where, \(CS_{ij}\) represents the categorized SHAP value for feature \(i\) in sample \(j\)

\subsubsection{SHAP Summary Dictionary}

Next, we initiate a SHAP summary dictionary (SSD) to capture the impact of features \(i\) for each class \(c\) in the dataset using Equation \ref{eq:shapSumm}.

\begin{equation}
      SSD = \{ c: \{ i: \{ CS: [ ]\}, ... \}, ... \}
\label{eq:shapSumm}
\end{equation}

where, \(CS\) represents \(\{P, N_{T}, N\}\)\\

\begin{figure*}[t]
    \centering
    \includegraphics[width=.6\linewidth]{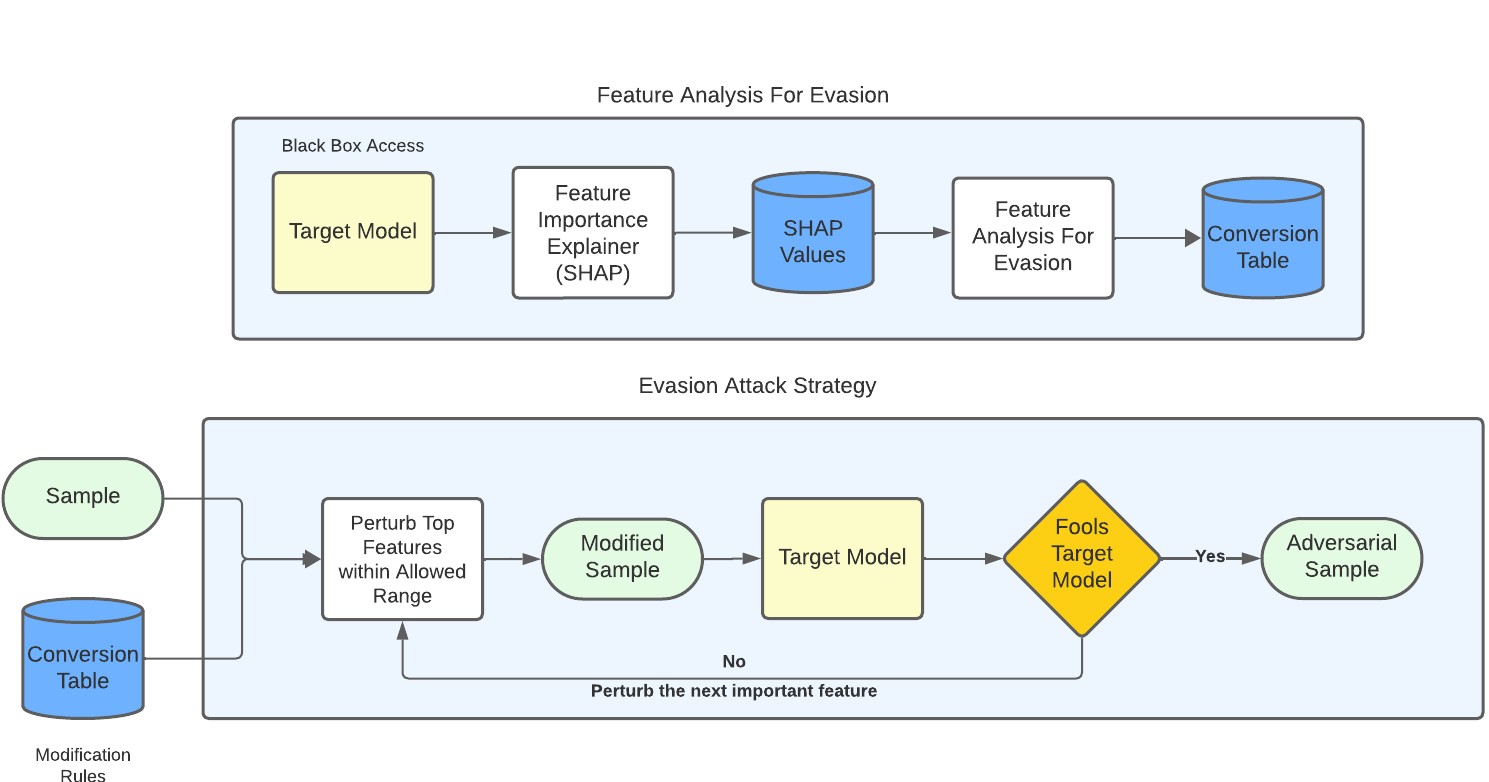}
    \caption{Feature Analysis For Evasion}
    \label{fig:faevasion}
\end{figure*}

Each class \( c \), feature \( i \), and SHAP category (CS) is represented, with the impact category (CF) accumulated in a list.

\subsubsection{Concise SHAP Summary Direction}

Using a count function, we quantify occurrences of impacts for each feature within each class. The sentiment (CS) with the maximum count for each impact (CF) and feature (i) in class (c) is then determined using Equations \ref{eq:consineShap} and \ref{eq:countimpact}.

\begin{equation}
      M_{c,i}(CF) = \arg\max_{CS} C_{i,j,CS}(CF)
\label{eq:consineShap}
\end{equation}
\begin{equation}
     C_{c,i,CS}(CF) = \text{\# of times } CF \text{ appears in } CS
\label{eq:countimpact}
\end{equation}

This aggregation process leads to a more concise SSD, enabling us to better understand the relationship between features and their impacts across different classes. Appendix \ref{sec:Appendix_A} Figure \ref{fig:concise_ssd} shows the Concise SSD for the Iris Dataset.

\subsubsection{Possible Class Conversions}

The set of possible class conversions, denoted as \textit{$class_{conversions}$}, is a set containing pairs $(i, j)$ where $i$ and $j$ are unique classes, and $i \neq j$. It includes all possible combinations of unique class pairs, ensuring that each pair consists of different classes, as defined by Equation \ref{eq:classconv}.
\begin{equation}
    \text{$class_{conversions}$} = \{(i, j) | i, j \in \text{classes } \text{and } i \neq j\}
\label{eq:classconv}
\end{equation}
% where, \text{class\_conversions} \hl {is a set containing pairs $(i, j)$ where $i$ and $j$ are unique classes, and $i \neq j$. In other words, it includes all possible combinations of unique class pairs, ensuring that each pair consists of different classes.}

\subsubsection{Feature Impact Aggregation}

To determine how feature impacts from one class $i$ can be converted to another $j$, an intersection method is used. This method involves intersecting the $negative$ and $neutral$ impacts for class $i$ $(I_i^-)$ with the $positive$ impacts for class $j$ $(positive_j)$. This intersection results in $P_{ij}$ which indicates a strong positive effect on class $j$ while exerting a negative impact on class $i$.

\textbf{Positive Effect on Class $j$:}
\begin{equation}
    P_{ij} = I_i^- \cap \text{positive}_j %\quad \text{strong \(positive\) effect on class \(j\)}
\label{eq:posImp}
\end{equation}

Conversely, the $positive$ and $neutral$ impacts for class $j$ $(I_j^+)$ are intersected with the $negative$ impacts for class $i$ $(negative_i)$. This intersection yields $N_{ij}$, signifying a strong negative effect on class $i$ with a positive effect on class $j$.

\textbf{Negative Effect on Class $i$:}
\begin{equation}
    N_{ij} = I_j^+ \cap \text{negative}_i %\quad \text{strong \(negative\) effect on class \(i\)}
\label{eq:negImp}
\end{equation}

The final step involves taking the union of $P_{ij}$ and $N_{ij}$. This union synthesizes the modifications necessary to promote a positive effect on class $j$ while simultaneously inducing a negative effect on class $i$.

\textbf{Final Effect to move from Class $i$ to Class $j$:}
\begin{equation}
    F_{ij} = P_{ij} \cup N_{ij}
    %\quad \text{strong \(negative\) effect on class \(i\)}
\label{eq:finalImp}
\end{equation}

Such modifications are critical in performing the targeted evasion attack, effectively moving from class $i$ to class $j$.

\subsubsection{Storing Conversion Directions}

In the final step of our process, we store conversion directions for each feature in the conversion table. These directions are pivotal in guiding modifications to feature values during a targeted evasion attack. Appendix \ref{sec:Appendix_A} Figure \ref{fig:conversion_table} shows the conversion table for the Iris Dataset.

% The culmination of this analysis is the creation of a conversion table. 
This table plays a pivotal role in mapping the impact of each feature from the original class to its potential impact on a target class, contingent upon modifications in feature values. The entire process of feature analysis for Evasion is encapsulated in the 'Feature Analysis Block' of Figure \ref{fig:faevasion}, providing a visual representation of the methodology and its components.
 
\subsection{Evasion Attack Strategy with Optimal Epsilon Technique}

\begin{figure}[t]
    \centering
    \includegraphics[width=.7\columnwidth]{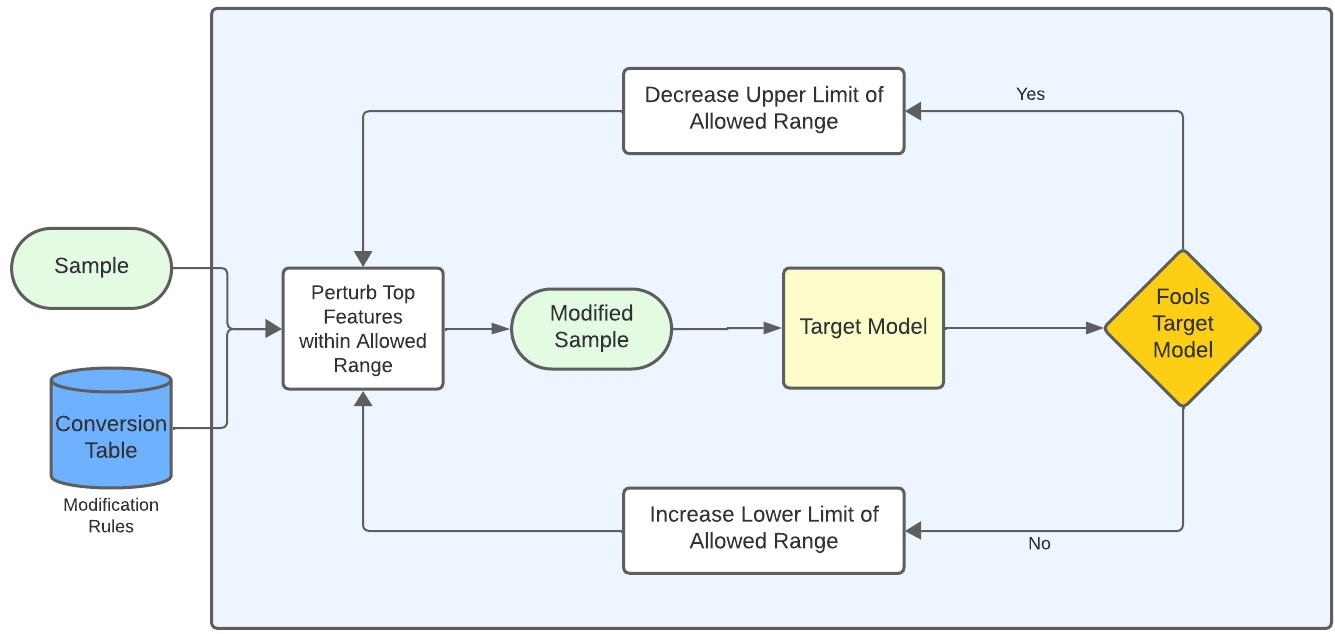}
    \caption{Optimal Epsilon}
    \label{fig:optepsilon}
\end{figure}

% \subsubsection{Evasion Attack Strategy with Optimal Epsilon Technique}
Our evasion attack strategy (refer Appendix \ref{sec:Appendix_B}, Algorithm \ref{ref:algoEvasion}) operates in a black-box setting, relying solely on the model's output predictions to guide the iterative process of modifying input features. The primary objective is to manipulate a given input sample, $x_{\text{org}}$, such that it deceives the machine learning model into misclassifying it as a target class, different from its original class.

This strategy leverages the knowledge acquired through the comprehensive feature importance analysis using SHAP, as detailed in Section 3.3 and Section 3.4. The conversion table, derived from this analysis, provides crucial insights into the directional adjustments required for each feature to facilitate the desired class conversion.

The evasion attack algorithm iterates over the features of the input sample, modifying them according to the conversion rules specified in the conversion table. The adjustments, denoted as $\Delta x_i$, are carefully calibrated to remain within plausible bounds, ensuring that the modifications do not exceed a predefined threshold, $d_{\text{max}}$, also referred to as the epsilon ($\epsilon$) value:

\begin{equation}
     x'_i = x_i + \Delta x_i
\label{eq:deltaAdj}
\end{equation}

During the attack process, we continuously monitor the distance between the modified sample, $x_{\text{adv\_temp}}$, and the original input, $x_{\text{org}}$, represented as $\text{distance}(x_{\text{adv\_temp}}, x_{\text{org}})$. The objective is to find the adversarial sample, $x_{\text{adv\_best}}$, that successfully triggers misclassification into the target class with minimal deviation from the original sample

\begin{equation}
    x_{\text{adv\_best}} = \arg\min_{x_{\text{adv\_temp}}} \text{distance}(x_{\text{adv\_temp}}, x_{\text{org}})
\label{eq:bestAdv}
\end{equation}

To refine the evasion attack approach and ensure the generation of effective adversarial samples with minimal perturbation, we introduce the Optimal Epsilon technique, as shown in Figure \ref{fig:optepsilon}. This technique systematically determines the smallest epsilon ($\epsilon_{\text{optimal}}$) necessary for successful evasion by employing a binary search loop, refer to Algorithm \ref{ref:algoOptimalEpsilon} in Appendix \ref{sec:Appendix_B}.

The binary search process starts with an initial epsilon range $[\epsilon_{\text{low}}, \epsilon_{\text{high}}] = [0, 0.5]$, where $0.5$ represents the upper limit for allowed perturbation. This upper limit can be extended to $1$ for determining the optimal epsilon across all samples without restrictions on perturbation magnitude. The loop iterates until the gap between $\epsilon_{\text{high}}$ and $\epsilon_{\text{low}}$ is less than a predefined $tolerance$ value.

Within each iteration, adversarial samples are generated by modifying feature values according to the conversion rules, aiming to shift the prediction from the original class to the target class. The effectiveness of these samples is evaluated on the target model. If a successful adversarial sample is found, the distance between the adversarial and original samples is calculated, and the optimal epsilon and the best adversarial sample are updated accordingly.

The binary search concludes once the difference between $\epsilon_{\text{high}}$ and $\epsilon_{\text{low}}$ falls below the $tolerance$ threshold. The final $\epsilon_{\text{optimal}}$ represents the minimum perturbation magnitude required for effective adversarial samples under the given conditions.

By integrating the Optimal Epsilon technique seamlessly into the evasion attack strategy, our methodology ensures the generation of precise and impactful adversarial samples, underscoring the vulnerability of machine learning models to carefully crafted evasion attacks.

% In summary, this technique refines the evasion attack strategy by applying a binary search to determine the least epsilon for effective adversarial sample generation, enhancing the precision and effectiveness of attacks on machine learning models.

\begin{figure*}[t]
\centering
\subfigure[SVM with RBF Kernel]{\includegraphics[width=2in]{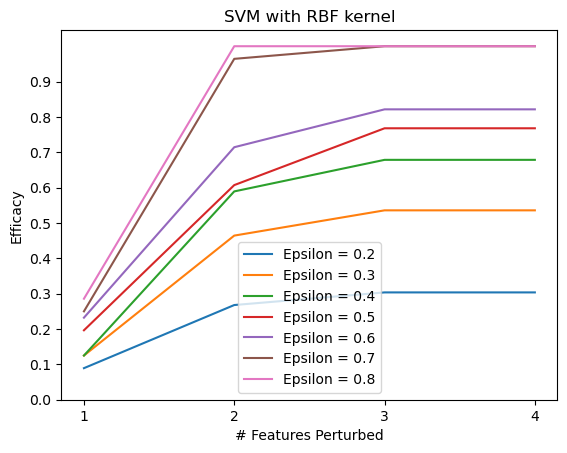}
\label{fig:mcsvmrbf}}
\hfil
\subfigure[SVM with Linear Kernel]{\includegraphics[width=2in]{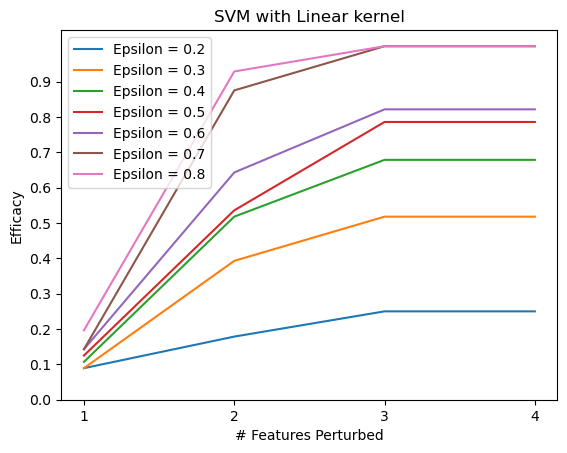}}
\label{fig:mcsvmlinear}
\hfil
\subfigure[XGBoost]{\includegraphics[width=2in]{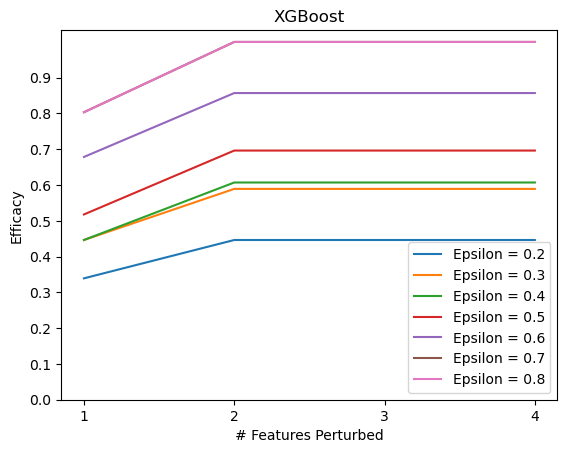}}
\label{fig:mcxgb}
\hfil
\subfigure[Decision Tree]{\includegraphics[width=2in]{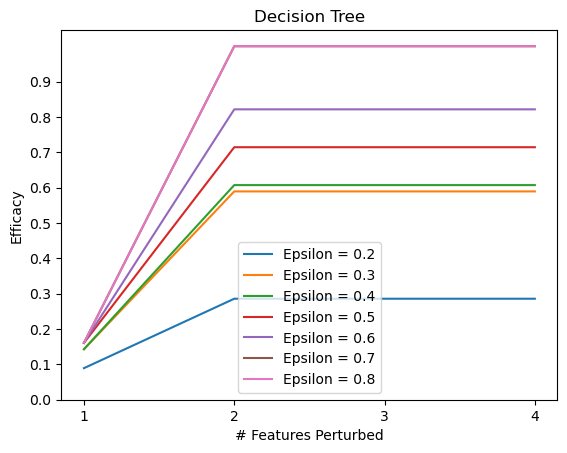}}
\label{fig:mcdt}
\hfil
\subfigure[Logistic Regression]{\includegraphics[width=2in]{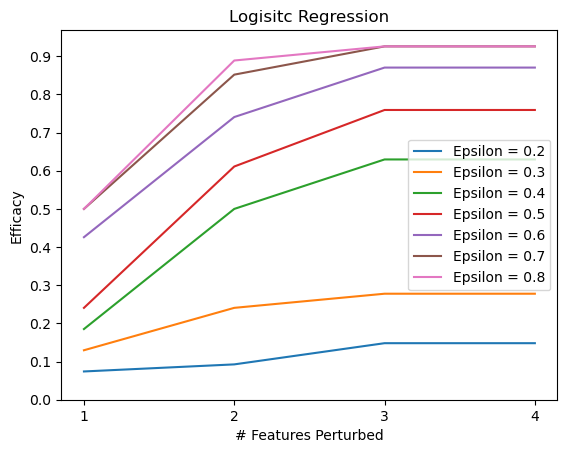}}
\label{fig:mclr}
       
\caption{Saturation Point: MultiClass Attacks on Iris Dataset}
     \label{fig:mcirisexperiments}
\end{figure*}

\section{Experiments}
\label{sec:Experiment}
\subsection{Efficacy as a Key Metric for Evasion}
Unlike traditional accuracy metrics in machine learning, efficacy in the context of our paper refers to the model's susceptibility to evasion attacks. Specifically, it measures the proportion of samples that successfully evade and deceive the model within the allowed perturbation limit, denoted as epsilon.

Let \(N\) be the total number of samples subjected to the evasion attack. Out of these, let \(N_{evaded}\) be the number of samples that successfully evade the model's detection within the perturbation limit. The efficacy, \(E\), can then be mathematically represented as:

\begin{table*}[t]
\renewcommand{\arraystretch}{1.3}
\setlength{\tabcolsep}{5pt}
% if using array.sty, it might be a good idea to tweak the value of
% \extrarowheight as needed to properly center the text within the cells
\caption{Efficacy of MultiClass Targeted Attack on Iris Dataset}
\label{tab:mctargeted}
\centering
% Some packages, such as MDW tools, offer better commands for making tables
% than the plain LaTeX2e tabular which is used here.
\begin{tabular}{|c||c|c|c||c|c|c||c|c|c||c|c|c||c|c|c|}
\hline
\multicolumn{1}{|c||}{\textbf{Epsilon }} & \multicolumn{3}{c||}{SVM (RBF)} & \multicolumn{3}{c||}{SVM (Linear)} & \multicolumn{3}{c||}{XGBoost} & \multicolumn{3}{c||}{Decision Tree} & \multicolumn{3}{c|}{Logistic Regression} \\
\hline
 $(\pmb{\epsilon})$ & \multicolumn{3}{c||}{Class} & \multicolumn{3}{c||}{Class} & \multicolumn{3}{c||}{Class} & \multicolumn{3}{c||}{Class} & \multicolumn{3}{c|}{Class} \\
% \cmidrule(lr){2-4} \cmidrule(lr){5-7} \cmidrule(lr){8-10}
% \hline
\hline
      & 0 & 1 & 2 & 0 & 1 & 2 & 0 & 1 & 2 & 0 & 1 & 2 & 0 & 1 & 2 \\
\hline
        \textbf{0.3} & 0.31 & 0.86 & 0.33 & 0.31 & 0.86 & 0.28 & 0.31 & 1 & 0.29 & 0.31 & 0.96 & 0.29 & 0.4 & 0.33 & 0.11 \\
\hline
        \textbf{0.4} & 0.62 & 1 & 0.4 & 0.625 & 1 & 0.33 & 0.37 & 1 & 0.29 & 0.37 & 1 & 0.29 & 0.73 & 0.95 & 0.16 \\    
\hline
        \textbf{0.5} & 0.94 & 1 & 0.4 & 1 & 1 & 0.33 & 0.69 & 1 & 0.29 & 0.69 & 1 & 0.29 & 1 & 1 & 0.28 \\
\hline
        \textbf{0.6} & 1 & 1 & 0.44 & 1 & 1 & 0.44 & 0.93 & 1 & 0.59 & 0.94 & 1 & 0.59 & 1 & 1 & 0.61\\
\hline
\end{tabular}
\end{table*}

\begin{equation}
% \begin{dmath}
     E = \frac{N_{evaded}}{N}
% \end{dmath}
\label{eq:deltafrac}
\end{equation}

A high efficacy score implies a greater number of samples evading the model successfully, indicating a potential vulnerability in the model's defense against adversarial attacks within the specified perturbation limit. By evaluating efficacy across different models and configurations, we can compare their robustness against evasion attacks, offering insights into the effectiveness of various defense mechanisms.

\subsection{Multiclass Classification - Iris Dataset}
In the context of multiclass classification, we extended our research to the Iris dataset \cite{misc_iris_53}, known for its suitability in multiclass tasks. We evaluated five machine learning architectures: SVM with RBF and Linear kernels, XGBoost, Logistic Regression, and Decision Tree \cite{Unwin2021TheID}.
Our focus was on perturbing features of original samples to generate adversarial samples, aiming at specific class misclassifications. In the multiclass setting, this approach introduced additional complexity due to the presence of multiple target classes. Table \ref{tab:mctargeted} illustrates the efficacy of targeted attacks across different machine learning models and varying epsilon values. Figure \ref{fig:mcirisexperiments} visualizes the relationship between the efficacy of attacks and the number of features perturbed. 
This provides deeper insights into the optimization of feature perturbations for successful evasion, showing how certain models reach a saturation point beyond which additional perturbations do not significantly increase the success of the attack.

\begin{figure*}[t]
\centering
\subfigure[SVM with Linear Kernel]{\includegraphics[width=.4\linewidth]{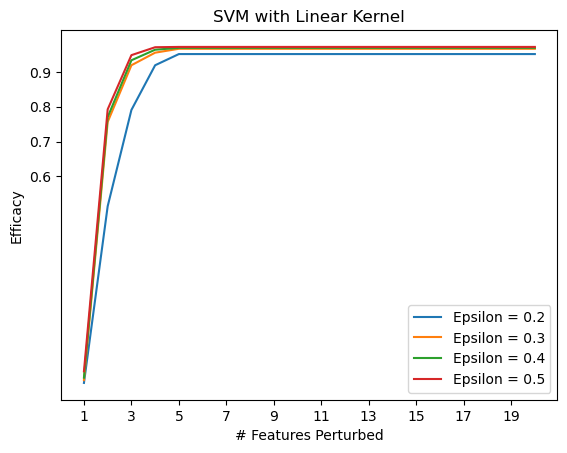}%
\label{fig:bcsvmlinear}}
\hfil
\subfigure[SVM with RBF Kernel]{\includegraphics[width=.4\linewidth]{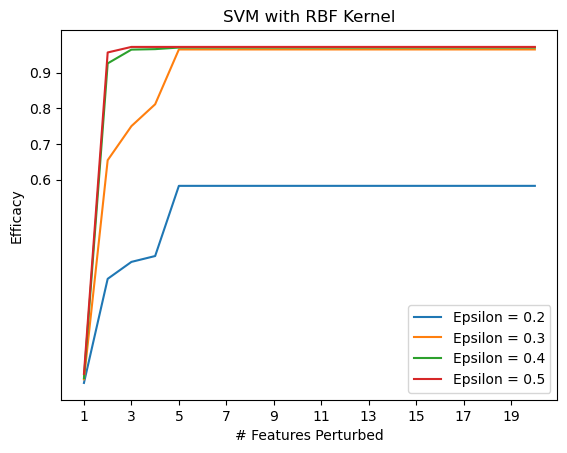}%
\label{fig:bcsvmrbf}}
\hfil
\subfigure[XGBoost]{\includegraphics[width=.4\linewidth]{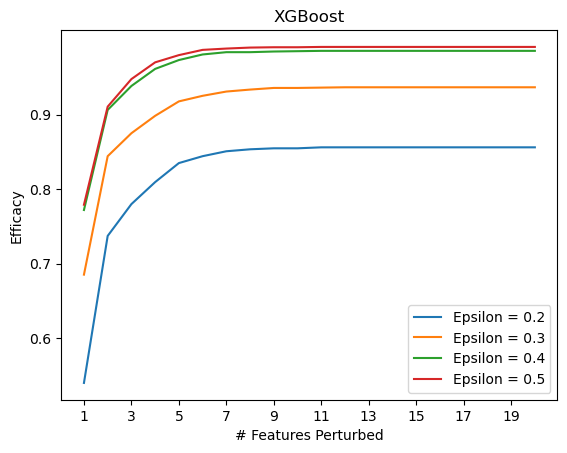}%
\label{fig:bcxgb}}
\hfil
\subfigure[Logistic Regression]{\includegraphics[width=.4\linewidth]{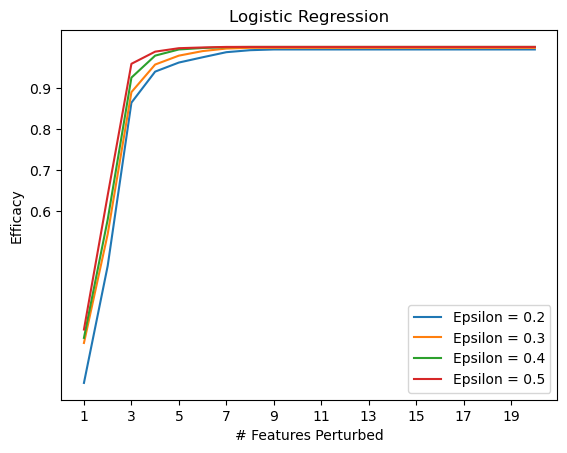}%
\label{fig:bclogistic}}
\caption{Saturation Point : Binary Class Attack on Bank Marketing Dataset}
\label{fig:bcexperiments}
\end{figure*}

\begin{table}[t]
\renewcommand{\arraystretch}{1.3}
% if using array.sty, it might be a good idea to tweak the value of
% \extrarowheight as needed to properly center the text within the cells
\caption{Efficacy of Binary Class Attack on Bank Marketing Dataset}
\label{tab:table}
\centering
% Some packages, such as MDW tools, offer better commands for making tables
% than the plain LaTeX2e tabular which is used here.
\begin{tabular}{|c||c||c||c||c|}
\hline
% One & Two\\
 \textbf{Epsilon} & SVM  & SVM & XGBoost &  Logistic \\
 $(\pmb{\epsilon})$ & (RBF) & (Linear) & & Regression \\
\hline
\textbf{0.2} & 0.58 & 0.95 & 0.86 & 0.99 \\
\hline
\textbf{0.3} & 0.96 & 0.96 & 0.94 & 1 \\
\hline
\textbf{0.4} & 0.97 & 0.97 & 0.98 & 1 \\
\hline
\textbf{0.5} & 0.97 & 0.97 & 0.99 & 1 \\
\hline
\end{tabular}
\end{table}

\begin{table*}[t]
\renewcommand{\arraystretch}{1.3}
\setlength{\tabcolsep}{2.2pt}
% if using array.sty, it might be a good idea to tweak the value of
% \extrarowheight as needed to properly center the text within the cells
\caption{Comparative Analysis of the Efficacy of Targeted Attacks on the SVM Model for Iris Dataset}
\label{tab:comparetargeted}
\centering
% Some packages, such as MDW tools, offer better commands for making tables
% than the plain LaTeX2e tabular which is used here.
\begin{tabular}{|c||c||c||c||c||c||c||c||c||c|}
\hline
\textbf{Epsilon} & \multicolumn{3}{c||}{Class 0} & \multicolumn{3}{c||}{Class 1} & \multicolumn{3}{c|}{Class 2} \\
% \cmidrule(lr){2-4} \cmidrule(lr){5-7} \cmidrule(lr){8-10}
\hline
      \textbf{$(\pmb{\epsilon})$} & FGM
 & PGD & MISLEAD & FGM & PGD & MISLEAD & FGM &  PGD & MISLEAD \\
\hline
        \textbf{0.2} & 0.02 & \textbf{0.04} & 0.03 & \textbf{0.58} & 0.22 & 0.35 & 0.28 & 0.25 & \textbf{0.4} \\
\hline
        \textbf{0.3} & 0.08 & 0.06 & \textbf{0.31} & 0.6 & 0.33 & \textbf{0.86} & 0.28 & 0.31 & \textbf{0.4} \\    
\hline
        \textbf{0.4} & 0.2 & 0.1 & \textbf{0.62} & 0.7 & 0.56 & \textbf{1} & 0.28 & 0.33 & \textbf{0.4} \\
\hline
\end{tabular}
\end{table*}
\subsection{Binary Classification - Bank Marketing Dataset}
In the binary classification context, experiments were carried out on a bank marketing dataset \cite{misc_bank_marketing_222}, a subset derived from the original Bank Marketing dataset from the UCI repository \cite{MORO201422}. These experiments yielded crucial insights into the robustness of various machine learning models and the dynamics of feature perturbations. We evaluated four machine learning architectures: SVM with Radial Basis Function (RBF) and Linear kernels \cite{10.1145/130385.130401}, XGBoost \cite{10.1145/2939672.2939785}, Logistic Regression.

% \begin{table}[t]
% % increase table row spacing, adjust to taste
% \renewcommand{\arraystretch}{1.1}
% % if using array.sty, it might be a good idea to tweak the value of
% % \extrarowheight as needed to properly center the text within the cells
% \caption{SVM Model Accuracy Before/After Evasion Attack on Iris Dataset}
% \label{tab:accuracyiris}
% \centering
% % Some packages, such as MDW tools, offer better commands for making tables
% % than the plain LaTeX2e tabular which is used here.
% \begin{tabular}{|c||c||c|}
% \hline
% % One & Two\\
%  \textbf{Epsilon $(\pmb{\epsilon})$} & Before  & After \\
% \hline
% \textbf{0.2} & 0.92 & 0.44 \\
% \hline
% \textbf{0.3} & 0.92 & 0.12 \\
% \hline
% \textbf{0.4} & 0.92 & 0.00 \\
% \hline
% \textbf{0.5} & 0.92 & 0.00 \\
% \hline
% \end{tabular}
% \end{table}

\begin{table}[t]
\renewcommand{\arraystretch}{1.1}
% if using array.sty, it might be a good idea to tweak the value of
% \extrarowheight as needed to properly center the text within the cells
\caption{Comparative Analysis of the Efficacy of Untargeted Attacks on the SVM Model for Iris Dataset}
\label{tab:compareuntargeted}
\centering
% Some packages, such as MDW tools, offer better commands for making tables
% than the plain LaTeX2e tabular which is used here.
\begin{tabular}{|c||c||c||c|}
\hline
% One & Two\\
 \textbf{Epsilon $(\pmb{\epsilon})$} & FGM  & PGD & MISLEAD \\
\hline
\textbf{0.2} & 0.34 & 0.52 & \textbf{0.53} \\
\hline
\textbf{0.3} & 0.48 & 0.68 & \textbf{0.86} \\
\hline
\textbf{0.4} & 0.78 & 0.79 & \textbf{1} \\
\hline
\end{tabular}
\end{table}

\begin{table}[t]
\renewcommand{\arraystretch}{1.3}
\setlength{\tabcolsep}{2.2pt}
% if using array.sty, it might be a good idea to tweak the value of
% \extrarowheight as needed to properly center the text within the cells
\caption{Impact of Evasion Attacks on Model Accuracy - Iris Dataset}
\label{tab:modelaccuracyiris}
\centering
% Some packages, such as MDW tools, offer better commands for making tables
% than the plain LaTeX2e tabular which is used here.
\begin{tabular}{|c||c|c||c|c||c|c|}
\hline
 \textbf{$(\pmb{\epsilon})$} & \multicolumn{2}{c||}{SVM [RBF]} & \multicolumn{2}{c||}{XGBoost} & \multicolumn{2}{c|}{Logistic Regression} \\
% \cmidrule(lr){2-4} \cmidrule(lr){5-7} \cmidrule(lr){8-10}
\hline
       & Before
 & After & Before & After & Before & After \\
\hline
        \textbf{0.2} & 0.92 & 0.44 & 0.96 & 0.48 & 0.94 & 0.56  \\
\hline
        \textbf{0.3} & 0.92 & 0.12 & 0.96 & 0.02 & 0.94 & 0.36 \\    
\hline
        \textbf{0.4} & 0.92 & 0.00 & 0.96 & 0.00 & 0.94 & 0.06 \\        
\hline
        \textbf{0.5} & 0.92 & 0.00 & 0.96 & 0.00 & 0.94 & 0.00 \\
\hline
\end{tabular}
\end{table}

\begin{table}[t]
\renewcommand{\arraystretch}{1.3}
\setlength{\tabcolsep}{2.2pt}
% if using array.sty, it might be a good idea to tweak the value of
% \extrarowheight as needed to properly center the text within the cells
\caption{Impact of Evasion Attacks on Model Accuracy - Bank Marketing Dataset}
\label{tab:modelaccuracybank}
\centering
% Some packages, such as MDW tools, offer better commands for making tables
% than the plain LaTeX2e tabular which is used here.
\begin{tabular}{|c||c|c||c|c||c|c|}
\hline
& \multicolumn{2}{c||}{SVM [RBF]} & \multicolumn{2}{c||}{XGBoost} & \multicolumn{2}{c|}{Logistic Regression} \\
% \cmidrule(lr){2-4} \cmidrule(lr){5-7} \cmidrule(lr){8-10}
\hline
      \textbf{ $(\pmb{\epsilon})$} & Before
 & After & Before & After & Before & After \\
\hline
        \textbf{0.2} & 0.91 & 0.38 & 0.96 & 0.16 & 0.91 & 0.04  \\
\hline
        \textbf{0.3} & 0.91 & 0.03 & 0.96 & 0.08 & 0.91 & 0.03 \\    
\hline
        \textbf{0.4} & 0.91 & 0.03 & 0.96 & 0.01 & 0.91 & 0.00 \\        
\hline
        \textbf{0.5} & 0.91 & 0.03 & 0.96 & 0.00 & 0.91 & 0.00 \\
\hline
\end{tabular}
\end{table}

Table \ref{tab:table} details the diverse performance spectrum across various models with increasing epsilon values. Notably, Logistic Regression exhibited high evasion susceptibility, reaching perfect evasion (efficacy of 1) at an epsilon value of 0.3 and maintaining this across higher epsilon values. In contrast, SVM with RBF kernel and XGBoost models demonstrated a gradual increase in their susceptibility to evasion attacks as epsilon increased, suggesting a more robust stance against smaller perturbations but a vulnerability at higher epsilon levels. Figure \ref{fig:bcexperiments} visually illustrates the saturation point in the number of features necessary for successful evasion attacks across different epsilon value. 
The saturation point is a critical concept, denoting the threshold beyond which increasing the number of perturbed features does not significantly enhance the success rate of the evasion attack.

\subsection{Comparative Study}
In our comparative study, we evaluate the performance of the MISLEAD technique against established adversarial defense methods, leveraging Fast Gradient Method (FGM) and Projected Gradient Descent (PGD) from Adversarial Robustness Toolbox (ART) \cite{art2018} and SecML \cite{melis2019secml} libraries.

For the Targeted Attack, experiments were conducted with each of the three Iris classes as the target, as shown in Table \ref{tab:comparetargeted}. MISLEAD consistently shows enhanced resilience against targeted attacks towards specific classes when compared to FGM and PGD.

Across various epsilon values in the Untargeted Attack, as shown in Table \ref{tab:compareuntargeted}, MISLEAD demonstrates competitive efficacy when compared to FGM and PGD. The qualitative analysis suggests that MISLEAD achieves robust results, especially at higher epsilon values, surpassing existing methods. Importantly, FGM and PGD are considered White Box attacks, while MISLEAD operates as a Black Box attack. This distinction adds a critical layer to the comparative analysis, as MISLEAD's efficacy under limited information about the model internals is a significant aspect in real-world scenarios.

These results collectively suggest that the MISLEAD technique, as a novel approach, displays promising performance in both attack scenarios. The qualitative insights emphasize its potential in providing robust adversarial defense, showcasing its superiority over existing methods across various attack scenarios on tabular data.

\subsection{Assessment on Model Accuracy}
Upon applying our evasion attack method, as detailed in Tables \ref{tab:modelaccuracyiris} and \ref{tab:modelaccuracybank}, we observe a decrease in accuracy models. For the iris dataset, accuracy drops from a stable 0.92 down to 0.00 with progressive increases in the perturbation $\epsilon$. This showcases the attack's capacity to significantly disrupt model performance. The bank marketing dataset shows a similar pattern, with accuracy falling from 0.96 to 0.00. These findings demonstrate our method's ability to highlight the vulnerabilities of machine learning models and stress the need for enhanced defensive measures.

While the current study focuses on tabular data, we believe the MISLEAD methodology can be extended to other data domains, such as images and audio, by leveraging appropriate model explanation techniques. For image data, methods like Grad-CAM \cite{8237336}, DeepLIFT \cite{10.5555/3305890.3306006}, and SHAP for images can provide explainable feature representations. Similarly, for audio data, techniques like Layer-wise Relevance Propagation \cite{Montavon2019} can extract interpretable features. By integrating these domain-specific feature importance analysis tools, the MISLEAD approach can identify vulnerabilities and generate targeted adversarial samples across diverse data modalities. This extensibility underscores the generalizability of the proposed methodology, fostering its applicability in securing machine learning systems handling various data types.

\section{Mitigation Strategies}
Mitigating evasion attacks on AI models is an active area of research, with several promising approaches. Adversarial training, as described by Madry et al. in \cite{madry2017towards}, exposes the model to both clean and adversarially crafted data, improving its robustness to slight variations used in evasion attempts. Defensive distillation, proposed by Goldblum et al. in \cite{goldblum2020adversarially}, leverages a pre-trained, robust model to train a new model to inherit that robustness. These techniques can be complemented by ensuring high-quality training data with inherent variations and employing ensemble methods for a more robust overall system. Continuous monitoring and adaptation to evolving threats remain essential for maintaining a strong defense.

\section{Conclusion}
\label{sec:Conclusion}
In this work, we have introduced a groundbreaking methodology that combines SHAP-based feature importance analysis with an innovative optimal epsilon technique, significantly amplifying the effectiveness of evasion attacks on machine learning models. This methodology is distinct in its capability to precisely identify and manipulate the most influential features of a learning model, thereby refining the accuracy of adversarial sample generation. Our study's cornerstone, the optimal epsilon technique, determines the minimal perturbation required for successful evasion, optimizing the evasion process and establishing a new benchmark in adversarial attack precision. Employing the SHAP framework, our approach not only deepens the understanding of model vulnerabilities but also facilitates the creation of targeted and highly effective adversarial samples, marking a novel advancement in the field.

Looking towards the future, several research avenues present themselves. One promising direction is the extension of our techniques to different data forms, including image and audio data, to comprehensively assess the vulnerabilities across various machine learning models. Additionally, applying our methods to advanced machine learning architectures, especially deep learning models, could provide invaluable insights into both offensive and defensive strategies in model security. Moreover, the development of robust defense mechanisms against the sophisticated evasion attacks demonstrated in our work stands as a critical area of future exploration. Through these endeavors, we aim to contribute to the ongoing advancement in the security of machine learning systems against increasingly sophisticated adversarial threats.

\bibliographystyle{unsrtnat}
\bibliography{references}  %%% Uncomment this line and comment out the ``thebibliography'' section below to use the external .bib file (using bibtex) .

% %%% Uncomment this section and comment out the \bibliography{references} line above to use inline references.
% % \begin{thebibliography}{1}

% % 	\bibitem{kour2014real}
% % 	George Kour and Raid Saabne.
% % 	\newblock Real-time segmentation of on-line handwritten arabic script.
% % 	\newblock In {\em Frontiers in Handwriting Recognition (ICFHR), 2014 14th
% % 			International Conference on}, pages 417--422. IEEE, 2014.

% % 	\bibitem{kour2014fast}
% % 	George Kour and Raid Saabne.
% % 	\newblock Fast classification of handwritten on-line arabic characters.
% % 	\newblock In {\em Soft Computing and Pattern Recognition (SoCPaR), 2014 6th
% % 			International Conference of}, pages 312--318. IEEE, 2014.

% % 	\bibitem{hadash2018estimate}
% % 	Guy Hadash, Einat Kermany, Boaz Carmeli, Ofer Lavi, George Kour, and Alon
% % 	Jacovi.
% % 	\newblock Estimate and replace: A novel approach to integrating deep neural
% % 	networks with existing applications.
% % 	\newblock {\em arXiv preprint arXiv:1804.09028}, 2018.

% % \end{thebibliography}

\onecolumn
\appendix
\newpage
\section{Concise SSD and Conversion Table}
\label{sec:Appendix_A}
In this section we present the compact SSD and the conversion table for the Iris Dataset. In this representation, the keys denote class conversions, while the corresponding list values illustrate the directional adjustments needed for each of the four features in the sample.
\begin{figure}[h]
    \centering
        \subfigure[Concise SSD For Iris Dataset]{\includegraphics[width=3.5in]{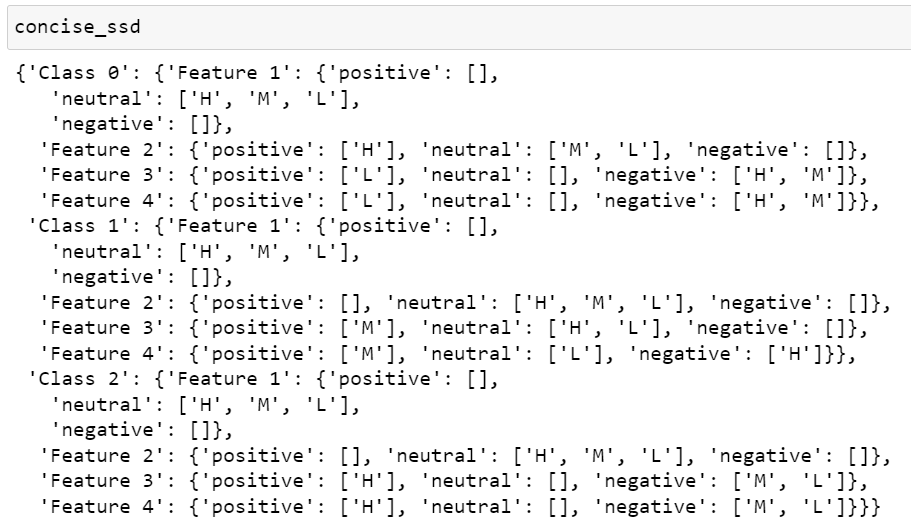}%
        \label{fig:concise_ssd}}
        \hfill
        \subfigure[Conversion Table For Iris Dataset]{\includegraphics[width=3.5in]{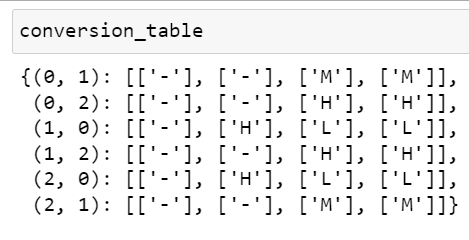}%
                \label{fig:conversion_table}}
            \caption{Concise SSD and Conversion Table}
        \label{fig:conv_table}

\end{figure}     

The feature analysis for evasion begins with the categorization of features based on their influence levels - Low (L), Medium (M), and High (H), as per Equation \ref{eq:festCat}. Following this, we progress to the initialization of the SHAP Summary based on Equation \ref{eq:shapSumm}, which involves segregating SHAP values into three distinct categories: positive $(P)$,  neutral $(N_{T})$, and negative $(N)$ based on the criteria detailed in Equation \ref{eq:shapCat}. This classification is pivotal in understanding the directional influence of each feature.

Once the SHAP summary is constructed, the subsequent step involves generating a concise summary. This is achieved by iterating through each class and feature, allowing us to categorize the impact of features within each class. To accomplish this, we analyze the occurrence counts of impact categories for each class and feature, considering the occurrence of these impact categories within the SHAP categories (Equation \ref{eq:consineShap} and \ref{eq:countimpact}).
Illustrating the process with an example: for a specific class and feature, if the impact category 'H' occurs 4 times in the 'positive' SHAP category and 2 times in the 'negative' SHAP category, we assign the impact category 'H' to the 'positive' category in the refined summary, prioritizing the higher occurrence within that particular SHAP category. This approach provides a balanced representation of feature impacts by averaging effects across the entire dataset, fostering a more robust understanding of relationships between feature values and their corresponding impact categories.

In addition to analyzing feature impact, we create a conversion table mapping the impact of each feature on the original class to its potential impact on the target class given a change in feature value. Initially, an empty conversion table is initialized, and possible class conversions are determined based on the number of unique classes present in the dataset (Equation \ref{eq:classconv}). For each class conversion, we iterate through features and consider impact categories within both the original and target classes. By comparing these impact categories, we identify the direction in which the feature’s value should be modified (Equation \ref{eq:posImp}, \ref{eq:negImp} and \ref{eq:finalImp}).

\newpage
\section{Algorithms}
\label{sec:Appendix_B}
\subsection{Evasion attack}
In this section we provide a detailed description of the evasion attack algorithm employed in our research. This algorithm operates in a black-box setting, relying solely on the model's output predictions to guide the strategy. The iterative process aims to modify the features of the input sample, ultimately crossing the decision boundary into the target class.

\begin{algorithm*}[h]
\caption{Evasion Attack Strategy}
\label{ref:algoEvasion}
\begin{algorithmic}
\STATE {\bfseries Input:} $x_{org}$, $c_{from}$, $c_{to}$
\STATE {\bfseries Output:} $x_{adv}$, $success$
\STATE {\bfseries Data:} $target_{model}$, $d_{max}$, $conversion_{table}$, $T_{low}$, $T_{high}$
\STATE $x_{adv} \leftarrow \text{copy of } x_{org}$
\STATE $best_{adv} \leftarrow \text{copy of } x_{org}$
\STATE $d_{least} \leftarrow 1$
\STATE $conversion_{rules} \leftarrow conversion_{table}[(c_{from}, c_{to})]$
\FORALL{ feature in $x_{org}$}
    \STATE Categorize $feature_{val}$ based on $T_{low}$ and $T_{high}$
    \FORALL{ $conversion_{category}$ in $conversion_{rules}$[feature]}
        \STATE $x_{temp} \leftarrow \text{copy of } x_{adv}$
        \IF{$conversion_{category}$ is '-' or the category equals $conversion_{category}$}
            \STATE \textbf{continue}
        \ENDIF
        \STATE Modify $x_{adv}$ from category to $conversion_{category}$
        \STATE Clip modified feature value to [0, 1]
        \IF{$target_{model}$.predict($x_{adv}$) == $c_{to}$}
            \STATE $d_{new} = \text{distance}(x_{adv}, x_{org})$
            \IF{$d_{new} < d_{least}$}
                \STATE $best_{adv} \leftarrow \text{copy of } x_{adv}$
                \STATE $d_{least} \leftarrow d_{new}$
            \ENDIF
        \ENDIF
    \ENDFOR
\ENDFOR
\IF{$best_{adv} \neq x_{org}$}
    \STATE \textbf{return} $best_{adv}$, \textbf{True}
\ELSE
    \STATE \textbf{return} $x_{adv}$, \textbf{False}
\ENDIF
\end{algorithmic}
\end{algorithm*}

\newpage
\subsection{Optimal Epsilon}
In this section we provide an algorithm to obtain an Optimal Epsilon, that determines the smallest epsilon, (\( \epsilon_{\text{optimal}} \)) required for creating impactful adversarial samples. It employs a refined evasion attack approach, utilizing a binary search loop to iteratively narrow down epsilon ranges, ensuring the generation of effective adversarial samples with minimal perturbation.

\begin{algorithm*}[h]
\caption{Optimal Epsilon}
\label{ref:algoOptimalEpsilon}
\begin{algorithmic}
\STATE {\bfseries Input:} $x_{\text{org}}$, $c_{from}$, $c_{to}$
\STATE {\bfseries Output:} $x_{\text{adv}}$, $d_{least}$, $\epsilon_{\text{optimal}}$, $success$
\STATE {\bfseries Data:} $target_{model}$, $conversion_{table}$, $T_{low}$, $T_{high}$, $tolerance$
\STATE $\epsilon_{\text{low}} \leftarrow 0$
\STATE $\epsilon_{\text{high}} \leftarrow 0.5$
\STATE $x_{\text{adv}} \leftarrow \text{copy of } x_{\text{org}}$
\STATE $best_{\text{adv}} \leftarrow \text{copy of } x_{\text{org}}$
\STATE $d_{least} \leftarrow 1$
\STATE $conversion_{rules} \leftarrow conversion_{table}[(c_{from}, c_{to})]$
\STATE $\epsilon_{\text{optimal}} \leftarrow 1$
\WHILE{$(\epsilon_{\text{high}} - \epsilon_{\text{low}}) > \text{tolerance}$}
    \STATE $\epsilon_{\text{mid}} \leftarrow (\epsilon_{\text{low}} + \epsilon_{\text{high}}) / 2$
    \STATE $x_{\text{adv}} \leftarrow \text{copy of } x_{\text{org}}$
    \FORALL{ feature in $x_{\text{org}}$}
        \STATE Categorize $feature_{val}$ based on $T_{low}$ and $T_{high}$
        \FORALL{ $conversion_{category}$ in $conversion_{rules}[feature]$}
            \IF{$conversion_{category}$ is '-' or the category equals $conversion_{category}$}
                \STATE \textbf{continue}
            \ENDIF
            \STATE Modify $x_{\text{adv}}$ from category to $conversion_{category}$
            \STATE Clip modified feature value to [0, 1]
            \IF{$target_{model}$.predict($x_{\text{adv}})$ == $c_{to}$}
                \STATE $d_{new} = \text{distance}(x_{\text{adv}}, x_{\text{org}})$
                \IF{$d_{new} < d_{least}$}
                    \STATE $best_{\text{adv}} \leftarrow \text{copy of } x_{\text{adv}}$
                    \STATE $d_{least} \leftarrow d_{new}$
                    \STATE $\epsilon_{\text{optimal}} \leftarrow \epsilon_{\text{mid}}$
                \ENDIF
            \ENDIF
        \ENDFOR
    \ENDFOR
    \IF{$(best_{\text{adv}} \neq x_{\text{org}})$}
        \STATE $\epsilon_{\text{high}} \leftarrow \epsilon_{\text{mid}}$
    \ELSE
        \STATE $\epsilon_{\text{low}} \leftarrow \epsilon_{\text{mid}}$
    \ENDIF
\ENDWHILE
\IF{$(best_{\text{adv}} \neq x_{\text{org}})$}
    \STATE \textbf{return} $best_{\text{adv}}$, $d_{least}$, $\epsilon_{\text{optimal}}$, \textbf{True}
\ELSE
    \STATE \textbf{return} $x_{\text{adv}}$, $d_{least}$, $\epsilon_{\text{optimal}}$, \textbf{False}
\ENDIF
\end{algorithmic}
\end{algorithm*}

\end{document}